\documentclass[sigconf]{aamas} 

\usepackage{balance} 

\usepackage{amsmath}
\usepackage{multirow}
\usepackage{multicol}
\usepackage{bbm}
\usepackage{hhline}
\usepackage{algorithm}
\usepackage{amsfonts}       
\usepackage[noend]{algpseudocode}
\usepackage{xcolor}

\setcopyright{ifaamas}
\acmConference[AAMAS '21]{Proc.\@ of the 20th International Conference on Autonomous Agents and Multiagent Systems (AAMAS 2021)}{May 3--7, 2021}{London, UK}{U.~Endriss, A.~Now\'{e}, F.~Dignum, A.~Lomuscio (eds.)}
\copyrightyear{2021}
\acmYear{2021}
\acmDOI{}
\acmPrice{}
\acmISBN{}

\acmSubmissionID{361}

\title[AAMAS-2021 Formatting Instructions]{Cooperative Policy Learning with Pre-trained Heterogeneous Observation Representations}

\author{Wenlei Shi}
\affiliation{
  \institution{Microsoft Research}
  \city{Beijing}
  \country{China}}
\email{Wenlei.Shi@microsoft.com}

\author{Xinran Wei}
\affiliation{
  \institution{Beijing University of Posts and Telecommunications}
  \city{Beijing}
  \country{China}}
\email{weixinran@bupt.edu.cn}

\author{Jia Zhang}
\affiliation{
  \institution{Microsoft Research}
  \city{Beijing}
  \country{China}}
\email{Jia.Zhang@microsoft.com}

\author{Xiaoyuan Ni}
\affiliation{
  \institution{Hong Kong University of Science and Technology}
  \city{Hong Kong}
  \country{China}}
\email{xniac@connect.ust.hk}

\author{Arthur Jiang}
\affiliation{
  \institution{Microsoft ARD Incubation Team}
  \city{Beijing}
  \country{China}}
\email{Shujia.Jiang@microsoft.com}

\author{Jiang Bian}
\affiliation{
  \institution{Microsoft Research}
  \city{Beijing}
  \country{China}}
\email{Jiang.Bian@microsoft.com}

\author{Tie-Yan Liu}
\affiliation{
  \institution{Microsoft Research}
  \city{Beijing}
  \country{China}}
\email{Tie-Yan.Liu@microsoft.com}

\begin{abstract}

Multi-agent reinforcement learning (MARL) has been increasingly explored to learn the cooperative policy towards maximizing a certain global reward. 
Many existing studies take advantage of graph neural networks (GNN) in MARL to propagate critical collaborative information over the interaction graph, built upon inter-connected agents.
Nevertheless, the vanilla GNN approach yields substantial defects in dealing with complex real-world scenarios since the generic message passing mechanism is ineffective between heterogeneous vertices and, moreover, simple message aggregation functions are incapable of accurately modeling the combinational interactions from multiple neighbors. 
While adopting complex GNN models with more informative message passing and aggregation mechanisms can obviously benefit heterogeneous vertex representations and cooperative policy learning, it could, on the other hand, increase the training difficulty of MARL and demand more intense and direct reward signals compared to the original global reward. 
To address these challenges, we propose a new cooperative learning framework with pre-trained heterogeneous observation representations. Particularly, we employ an encoder-decoder based graph attention to learn the intricate interactions and heterogeneous representations that can be more easily leveraged by MARL. Moreover, we design a pre-training with local actor-critic algorithm to ease the difficulty in cooperative policy learning.
Extensive experiments over real-world scenarios demonstrate that our new approach can significantly outperform existing MARL baselines as well as operational research solutions that are widely-used in industry.

\end{abstract}


\newcommand{\BibTeX}{\rm B\kern-.05em{\sc i\kern-.025em b}\kern-.08em\TeX}

\begin{document}


\pagestyle{fancy}
\fancyhead{}


\maketitle 


\section{Introduction}
\label{intro}
Cooperative multi-agent systems (MAS) \citep{Panait2005} have been employed to deal with a wide range of real-world applications, such as large scale fleet management \citep{lin2018efficient}, empty container repositioning \citep{li2019cooperative}, and traffic signal control \citep{wei2019colight}.
Instead of following classical single-agent planning methods, cooperative MAS stands essentially as a more efficient paradigm to resolve the global optimization task through decentralized agents with a relatively-simplified agent-wise problem-solving scheme.
Essentially, the success of such cooperative MAS highly relies on its capability to model the cooperation between agents with the purpose of pursuing a maximized global utility.
In reality, to bolster cooperation among complex relationships between agents and environmental entities, it is indispensable for each agent to fully perceive the state information about other inter-correlated agents and environmental entities. 

To this end, previous studies~\citep{jiang2018graph, malysheva2019magnet} proposed to model the entire environment with an \emph{interaction graph} where vertices represent the agents and environmental entities and edges exist when two vertices can interact with each other. And, they leveraged Graph Neural Networks (GNN)~\citep{kipf2016semi,velivckovic2017graph,NIPS2017_6703} to propagate critical cooperation signals between inter-connected agents within the Multi-Agent Reinforcement Learning (MARL) framework~\citep{hu1998multiagent}.
However, while the vanilla GNN model can perform well in relatively simple games, two major weaknesses limit its representability to capture the state information through the interaction graph in many complex real-world scenarios. 
First, the generic procedure of message propagation in vanilla GNN assumes that the vertices in the interaction graph are homogeneous. 
In practice, however, the vertices can be heterogeneous, which impedes the efficient message passing 
due to the incompatible feature spaces between heterogeneous vertices. 
Furthermore, another weakness lies in simply using pooling function, such as average and maximization, as message aggregation, which may cause significant information loss or the over-smoothing problem. 
Although more expressive pooling functions such as LSTM have been employed to enhance GNN structures~\cite{NIPS2017_6703}, none of them, to our best knowledge, has been applied in MARL scenarios.

Henceforth, it becomes necessary to design a more expressive GNN with more informative message passing and aggregation mechanisms to increase the individual agent’s ability to grasp the complex representations of neighbors’ messages. 
Such enhancement over GNN, on the other hand, is likely to encounter a dilemma. 
Specifically, it may lead to drastically-soaring training difficulties, in terms of requiring either stronger training signals or much more experience, for each agent to converge into an effective cooperative policy. 
{However, the signals measuring the accomplishment of the global optimization goals, e.g. the total value of served orders in a ride-sharing platform\cite{lin2018efficient} or total shortage of empty containers in ocean transportation~\cite{li2019cooperative}, are influenced by the entangled actions from multiple agents and one agent could hardly evaluate its own actions with limited experiences especially when other agents' policy is also non-stationary. Merely leveraging such global signals and letting agents learn \emph{from scratch} will lead to poor sample efficiency and make the training process unbearably slow.}
To obtain effective information representation from the heterogeneous interaction graph and learn the cooperative policy for decentralized agents, in this paper, we propose a new multi-agent reinforcement learning framework. Particularly, this framework leverages the graph attention with a delicately designed encoder-decoder structure for efficient message passing and aggregation. Moreover, we introduce a pre-training technology to alleviate the difficulty of learning the cooperative policy with global rewards.


More concretely, to magnify the GNN's representability in the process of information passing and aggregation, we employ an encoder-decoder based graph attention network (EncGAT). 
Similar to its previous success in the NLP domain, this approach can
consecutively embed a sequence of neighboring vertices into a high-level representation space with a self-attention network and then generate the aggregated feature through a decoder attention network with the feature of the vertex itself as the query, the details of which are demonstrated in Figure \ref{fig:eda}. 

To ease the difficulty in learning cooperative policy for decentralized agents with global rewards, we propose a pre-training with a local actor-critic (PreLAC) algorithm. Specifically, a pre-training phase is designed to learn a selfish policy for each agent by using local rewards. Since local rewards, which are naturally existing in most global optimization scenarios (e.g. regional revenue of ride-sharing~\cite{lin2018efficient} and port-level container shortage \cite{li2019cooperative}), are mainly affected by the corresponding individual agent and its neighbors, it is much easier to train a selfish policy, specified by a local actor and a local critic with the shared EncGAT layer, for each agent. Following this pre-training, a formal training phase will be carried out, in which decentralized agents with cooperative local actors and a shared global critic will be continuously trained, starting from using the pre-trained EncGAT layer, towards maximizing the global rewards. To avoid parameter interference, local critics are still working in the formal training phase for the parameters updating in EncGAT.
Compared to the straightforward approach with a simple combination of global and local rewards, the PreLAC algorithm does not need to manually trade off between convergence and optimality, which can thus result in a more effective cooperative policy with less manual intervention.

To evaluate the effectiveness of our framework, we conduct experiments over a real-world problem, i.e., empty container repositioning, in the scenario of ocean transportation. Extensive experiments have shown that our new framework can significantly outperform either the operations research solution, a widely used industry standard, or existing GNN-powered MARL baselines. Further investigations have also been conducted to demonstrate the advantages of the proposed encoder-decoder attention structure and the introduction of the pre-training task in terms of their effectiveness in accelerating the convergence and improving the ultimate global performance.

Our main contribution can be summarized as follows:
\begin{itemize}
  \item We propose an encoder-decoder based graph attention (EncGAT) layer for representation learning of the complicate interactions between heterogeneous entities in the MAS with global optimization goal.
  \item We introduce the pre-training with local actor-critic (PreLAC) algorithm for cooperative policy learning, in which the selfish policy learning with the purpose of local rewards maximization is taken as a pre-trained task and the pre-trained EncGAT parameters will be used for further cooperative policy learning.
  \item Extensive experiments are conducted on the practical empty container repositioning (ECR) problem and the results demonstrate the advantages of the proposed framework.
\end{itemize}
In addition to the ECR problem, the proposed EncGAT-PreLAC framework in fact can be applied to a series of global optimization scenarios, where local rewards can be defined. These scenarios include but are not limited to vehicle re-balancing in ride-sharing, traffic control, and even football match.


\section{Related Work}

Recent years have witnessed the increasing research attention on applying cooperative multi-agent systems into many application scenarios.  
Previously, traditional operational research (OR) methods were widely used to achieve global optimization in a centralized way~\citep{epstein2012strategic, song2009empty, SONG20121556}. However, the prerequisite of both the forecasting module and artificial constructions of mathematical objectives and constraints gave rise to unsatisfactory performance in practice since OR methods are difficult to be adapted to a dynamic environment. 
With the recent rapid development of reinforcement learning, a rising number of efforts have turned to formulate MAS into stochastic games and then solve them using the multi-agent reinforcement learning (MARL) approach. There have been a couple of corresponding practices, such as empty container repositioning~\citep{li2019cooperative}, shared bike repositioning~\citep{li2018dynamic, pan2019deep, ghosh2017dynamic}, ride-sharing fleet management~\citep{fagnant2018dynamic, lin2018efficient}, traffic light control~\citep{wei2019colight}, and cloud resource allocation~\citep{wu2011novel, liu2017hierarchical}.

Yet, in many real-world MAS with complex networks, one critical challenge, particularly in terms of how to feed each agent with the informative representation of a comprehensive observation scope for its decision-maker, has not been fully addressed. Some previous studies simply produced such representation using pre-defined contextual statistics, which has to be coupled with specific scenarios and apparently limits the generality. Most recently, GNN has been increasingly used to learn and enhance the representation of observation scope in the non-Euclidean space. 
For instance, MAGnet~\citep{malysheva2019magnet} learns relevance information from the observation in the form of a relevance graph, where relation weights are learned by pre-defined loss function based on heuristic rules. 
The relation weights in DGN~\citep{jiang2018graph} are learned with a temporal regularization under the assumption that the attention of one agent does not change much in two adjacent time steps, which is not always true in practical problems yet. Another work~\citep{agarwal2019learning} leverages the inductiveness of graph attention to learn a transferable policy.
There have also been recent efforts to use GNN to enhance collaboration in a practical problem called traffic signal control~\citep{wei2019colight,devailly2020igrl}.

However, all the studies above treat the observation as an isomorphic graph, ignoring the heterogeneity of agents and environmental entities commonly existed in real-world scenarios.
A most recent work~\citep{sun2020scaling} exploits the sparsity of interactions between agents and tackles the problem of heterogeneity of agents from different teams. Other than ~\citep{sun2020scaling}, in this paper, we focus on the gap of feature space between both agents and environmental entities and design a new graph attention based aggregation approach.

There are also some existing studies paying attention to enhancing the training signal. Counterfactual advantage~\citep{AAAI1817193} is used for credit assignment to filtering out contributions of other agents. Cooperative reward mechanism design~\citep{li2019cooperative} is used to compute the cooperative local reward. Value decomposition~\citep{sunehag2018value, rashid2018qmix} is also a hot topic to solve the problem of cooperative MARL with a single joint reward signal. In this paper, we leverage the direct local reward to accelerate the training, but their methods can be integrated into our framework.

In the field of GNN, our work can be viewed as a graph-level training task with vertex-level pre-training. Recently, pre-training GNNs in supervised or semi-supervised tasks have also been studied in several works\citep{Hu*2020Strategies, navarin2018pretraining}, but our pre-training pipeline is quite different from theirs. In addition, there are several studies\citep{wang2019heterogeneous} about the heterogeneous graph. However, our proposed networks introduce a different encoder-decoder structured aggregation function. Besides, our work aims to solve the sequential decision-making task, which basically yields a different problem domain other than theirs. 

\section{Problem Formulation}

\paragraph{Heterogeneous interaction graph.} In many practical MAS, agents need to cooperate to achieve a maximized global utility. An interaction graph can be constructed to characterize the interaction between agents, e.g. fleets~\citep{lin2018efficient} and ports~\citep{li2019cooperative}, and environment entities, e.g. wooden walls~\citep{malysheva2019magnet}. Specifically, the vertices, in the interaction graph, are the agents and environmental entities, while edges stand for their corresponding interactions. Because agents and environmental entities usually yield different types, the interaction graph is essentially a heterogeneous graph. We follow the previous work \citep{sun2013mining} to formally define the heterogeneous graph.
\begin{definition} \label{def:hstg} A heterogeneous interaction graph, denoted as $\mathcal{G}={(\mathcal{V}, \mathcal{E})}$, is a graph including a set of vertices $\mathcal{V}$ and a set of directed edges $\mathcal{E}$, where $\mathcal{V}^* \subset \mathcal{V}$ is the set of agents. In $\mathcal{G}$, we denote the set of the types of vertices and edges as $\mathcal{\chi}$ and $\mathcal{\psi}$, and denote the type of vertex $i$ and edge $j$ as $\chi_i$ and $\psi_j$. The set of the types of edges that are connected to vertex $v_i$ are denoted as $\dot{\mathcal{N}}_{\psi}: v_i \rightarrow \mathbb{D}$.
Then we define the mapping function from the vertex to the set of neighbors that connected by the edges of type $d\in \dot{\mathcal{N}}_{\psi}(v_i)$ as $\mathcal{N}_d: v_i \rightarrow \mathbbm{V}$.

\end{definition}

\paragraph{MAS with heterogeneous interaction graph.} \label{def:semimdp}

We formally model the MAS with heterogeneous interaction graph as a Semi-POMDP $\left\langle \mathcal{V^*}, \mathcal{A}, \mathcal{S}, \mathcal{O}, \mathcal{R}, \mathcal{P}, \gamma \right\rangle$. 
More concretely, $\mathcal{V^*}$ denotes the aforementioned agent set;
$\mathcal{A}$ denotes the action set specified by the joint action space of all agents, i.e., $\mathcal{A} = \mathcal{A}_1\times \mathcal{A}_2 \times \cdots \times \mathcal{A}_{|\mathcal{V}^{*}|}$;
$\mathcal{S}$ represents the global state that can be defined by a heterogeneous interaction graph $\mathcal{G}$ comprised of the agent vertices, environmental entity vertices and interaction edges
; $\mathcal{O}$ represents the set of each agent's observations, which is the joint observation space, i.e., $\mathcal{O}=\mathcal{O}_1\times \mathcal{O}_2 \times \cdots \times \mathcal{O}_{|\mathcal{V}^*|}$, where $\mathcal{O}_i$ is the observation space of agent $i$ and contains the local vertices and interaction edges near the agent;
$\mathcal{R}$ is the global reward function, which is defined as $\mathcal{S} \times \mathcal{A} \rightarrow \mathbb{R}$; note that, in most of MAS, there exists local rewards for each agent\footnote{For examples, scores of a football player, orders completed by a driver and shortage of containers at a port.}, and we can similarly define the local reward function $R^{\textsc{loc}}_i: \mathcal{S}\times\mathcal{A}\rightarrow\mathbb{R}$;
$\mathcal{P}$ is the transition probability function $ \mathcal{S}_t \times \mathcal{A} \rightarrow \Delta\left(\mathcal{S}^{'}_{t+k}\right)$, where $\Delta$ is the probability simplex over the next global state and the time interval $k$ is non-constant time interval from current state to the next state;
$\gamma$ is the discount factor on the unit time interval
The policy of agent $i$ is defined as a function ${\pi_i}: O_i\rightarrow \mathcal{A}_i$. The joint policy is defined as $\boldsymbol{\pi}=\left(\pi_1,\cdots, \pi_{|\mathcal{V^*}|}\right)$.

\section{Proposed Framework}
\begin{figure*}[!]
  \centering
  \includegraphics[width=0.9\linewidth,trim= 0mm 2mm 0mm 2mm, clip]{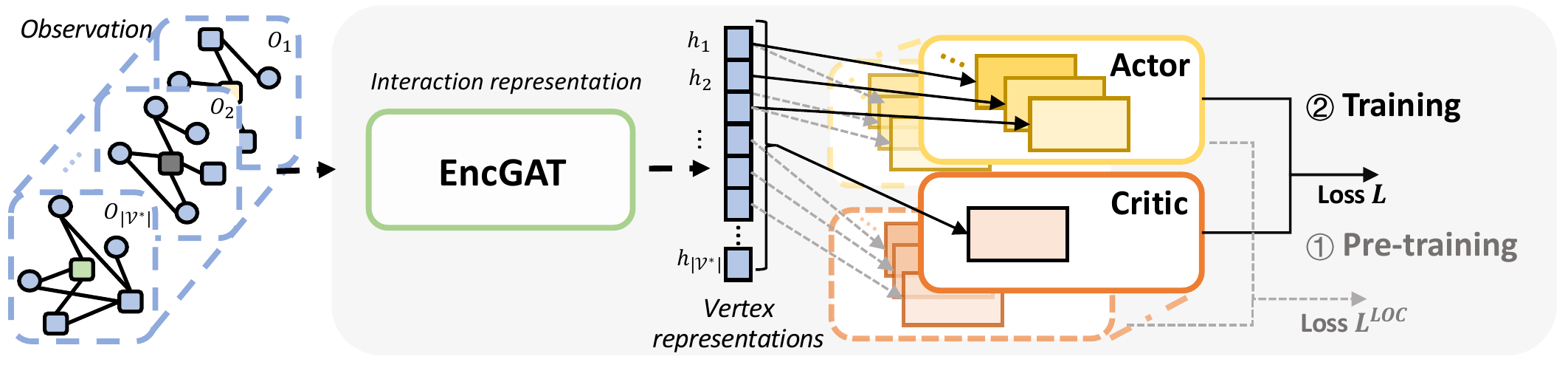}
  \caption{
  The overall structure of EncGAT-PreLAC. From left to right, it uses EncGAT for interaction representation and feeds the representations to the actor and critic headers. The overall model is first pre-trained with the local loss $L$ and then fine-tuned with a global loss $L^{\textsc{loc}}$ which are defined in Formula~\eqref{eqn:loc_losses} and Formula~\eqref{eqn:losses}.
  }
  \label{fig:model}
\end{figure*}

We introduce our framework based on the multi-agent A2C algorithm for convenience. In fact, our framework can be easily extended to other actor-critic based algorithms. As is shown in Figure~\ref{fig:model}, the proposed EncGAT-PreLAC framework is composed of two parts: an encoder-decoder graph attention (EncGAT) model and a pre-training procedure with local actor-critic (PreLAC). The EncGAT model is designed to learn the informative representation of the interaction graph, which will be fed into both the actor and critic networks to facilitate learning the cooperative policy. However, introducing the delicate-designed EncGAT model into the A2C algorithm makes the learning process more difficult in MAS. To improve both the learning efficiency and the effectiveness of the policy, we first leverage the local reward to train multiple selfish local actors and critics with the EncGAT model. Then, we initialize the final multiple actors and the global critic with the pre-trained EncGAT and conduct the fine-tuning with the global reward. More details are introduced in the following sub-sections.

\vspace{-2mm}

\subsection{Encoder-Decoder Based Graph Attention for State Representation}
\label{sec:eda_general}
As aforementioned, the vertices heterogeneity together with the intricate interactions between them prevents vanilla GNN to effectively learn the informative representations from the complex interaction graph. The EncGAT model is designed to solve these problems. 

\begin{figure}[!ht]
  \centering
  \includegraphics[width=0.95\linewidth,trim= 0mm 0mm 0mm 0mm, clip]{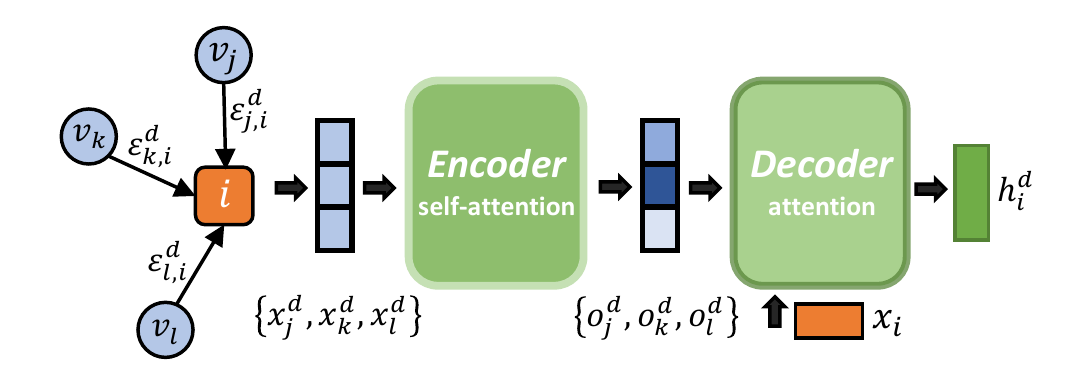}
  \caption{Encoder-decoder attention to aggregate interaction information from neighbors.}
  \label{fig:eda}
\end{figure}

The encoder-decoder based attention procedure is shown in the figure~\ref{fig:eda}. For vertex (agent) $v_i\in \mathcal{V^*}$ with feature vector $x_i$, its neighbor set with type $d$ is denoted as $\mathcal{N}_d(v_i)$ and the corresponding feature matrix is $\mathbf{x}_i^d$. The calculation procedure of the encoder is based on self-attention, which is similar to the transformer block and shown below:
\begin{equation}
\label{eqn:self-att}
\begin{aligned}
\mathbf{a}_i^d &= \textsc{Att}\left(W^{\mathcal{Q}}_{ec}\mathbf{x}_i^d, W^{\mathcal{K}}_{ec}\mathbf{x}_i^d, W^{\mathcal{V}}_{ec}\mathbf{x}_i^d\right) \\
\mathbf{k}_i^d &= \textsc{LN}\left(\mathbf{a}_i^d + \mathbf{x}_i^d\right) \\
\mathbf{z}_i^d &= \textsc{ReLU}\left(\mathbf{k}_i^d W_1^T + b_1\right)W_2^T + b_2\\
\mathbf{o}_i^d &= \textsc{LN}\left(\mathbf{k}_i^d + \mathbf{z}_i^d\right)
\end{aligned}
\end{equation}
where $\textsc{Att}$ is the scaled dot product attention which is the same as that in Transformer~\cite{NIPS2017_7181}, $W^{\mathcal{Q}}_{ec}$,$W^{\mathcal{K}}_{ec}$ and $W^{\mathcal{V}}_{ec}$ are the projection matrices of query, key and value in the attention function, $W_1$ and $W_2$ are the square matrices of the linear layers, $b_1$ and $b_2$ are the bias parameters, $\textsc{LN}$ is the layer norm and $\textsc{ReLU}$ is the ReLU activation function. The final embedding $o_i^d$ is already transformed into the same vector space with $x_i$ in case that $v_i$ is not of type $d$. We the whole encoder process as the function $\textsc{Enc}$. Through the function $\textsc{Enc}$, the feature of each neighboring vertex of $v_i$ is represented by considering other neighbors' information in the $\textsc{Att}$ function. 

The aggregation among the same type of neighbors is conducted through the decoder attention as is shown in the left part of the figure~\ref{fig:eda}. In decoder attention, another scaled dot product attention is applied, which uses the feature vector of $v_i$, i.e. $x_i$, as the query, and the encoded neighbors' features matrix $\mathbf{o_i^d}$  as the keys and values. To align the dimension, the feature vector $x_i$ is first expanded as a matrix $\mathbf{x_i}$.
The mathematical description is shown as follows:

\begin{equation}
\begin{aligned}
\mathbf{e}_i^d &= \textsc{Att}\left(W^{\mathcal{Q}}_{dc} \mathbf{x_i}, W^{\mathcal{K}}_{dc}\mathbf{o}_i^d, W^{\mathcal{V}}_{dc}\mathbf{o}_i^d\right) \\
\mathbf{f}_i^d &= \textsc{LN}\left(\mathbf{e}_i^d + \mathbf{o}_i^d\right) \\
\mathbf{c}_i^d &= \textsc{ReLU}\left(\mathbf{f}_i^d W_3^T + b_3\right)W_4^T + b_4\\
\mathbf{h}_i^d &= \textsc{LN}\left(\mathbf{f}_i^d + \mathbf{c}_i^d\right)
\end{aligned}
\end{equation}
where $W^{\mathcal{Q}}_{dc}$, $W^{\mathcal{Q}}_{dc}$ and $W^{\mathcal{Q}}_{dc}$ are the projection matrices of query, key and value in the attention function, $W_3$ and $W_4$ are the square matrices of linear layers and $b_3$ and $b_4$ are the bias parameters. Then we squeeze the expanded dimension of $\mathbf{h}_i$ into the vector $h_i$

Finally, as the neighbors of vertex $v_i$ may have different types, we simply concatenate all the output $h_i^d$ as the final feature of vertex $v_i$. The expression is as 
$h_i = \left \Vert {\left\{h_i^d | d \in \dot{\mathcal{N}}_{\psi}\left(v_i\right) \right\}}.\right.$

\subsubsection{Discussion on EncGAT} Indeed, we provide a new perspective on the cooperative information aggregation within a heterogeneous graph through EncGAT. 
To cooperate with other agents, an agent must perceive the heterogeneous information from neighbors. 
The success of applying the self-attention architecture in the natural language processing tasks inspires us to leverage the same idea to enforce cooperative information aggregation. 
For one thing, it enables the agent to parse the contextualized interactions with each individual neighbor given all other neighbors.
Moreover, the encoder-decoder attention can eliminate the feature gap between two heterogeneous vertices by mapping them into a unified representation space and reduce the difficulty in cooperative information extraction. 
The vanilla graph attention in fact can be seen as a simplified case without the encoder, which considers the interaction with each neighbor independently and then aggregates with a simple summation. This can only model the simplest linear combinational interactions of neighbors and will fail to model complex combinational interactions such as the relative order of consumer and producer in the inventory management, as is stated in Section~\ref{intro}. 
The self-attention we use in the encoder can be seen as weaving a complete graph between the neighbors of the same type and conducting the graph attention in these `sub-graphs'.
With the aid of self-attention in the encoder, the feature of each neighbor is calculated based on contextualized information from other neighbors such that it can capture the complex reactions in the interactions. 

In addition, the encoder-decoder structure we state here differs from the Transformer~\citep{NIPS2017_7181} in that the decoder in our model only gets a single vector as query, rather than a sequence. Therefore, no self-attention on the query is used in the decoder. The recent work GTR~\citep{li2019graph} uses transformer blocks in graph attention, which directly adds up features of the vertex and the weighted neighborhood in inter-graph message passing. It can be seen as the decoder-only version of EncGAT. 

\subsection{Pre-Training with Local Actor-Critic}
\label{sec:pretraining}
A crucial challenge of training the EncGAT lies in that a direct optimization towards the global reward may lead to insufficient observation understanding, which will slow down the convergence and even erode the performance of the learned cooperative policy. In this paper, we manage to address this problem by proposing the pre-training with the local actor-critic (PreLAC) algorithm. 

In the local actor-critic task, each agent is trained to maximize its own interest, that is the discounted sum of its local reward $\left\{r^{\textsc{loc}}_{i,t}\,|\, 1\leq i\leq |\mathcal{V}^*|\right\}$, where $t$ is the time-step. It encourages each agent to learn the concrete business logic about how the local reward is generated and how the neighborhood interactions affect the reward generation.

As is shown in figure~\ref{fig:model}, the EncGAT network is used as a shared embedding module for the actor and critic networks in the pre-training task. It receives the vertex-level training signals from the local actor and local critic headers. 
The Formula~\eqref{eqn:loc_losses} shows the training loss in this stage.
\begin{equation}
\label{eqn:loc_losses}
\begin{aligned}
\hat{A}_{i,t}^{\textsc{loc}} &= r_{i,{t+1}}^{\textsc{loc}} + \cdots + \gamma^{k-1}r_{i,{t+k}}^{\textsc{loc}} + \gamma^{k} {V^{\pi_i}}\left(O_{i, t+k}\right) - {V^{\pi_i}}\left(O_{i, t}\right),
\\
\delta_{i,t}^{\textsc{loc}} &= \left(\hat{A}_{i,t}^{\textsc{loc}}\right)^2, 
\\
\phi_{i,t}^{\textsc{loc}} &= - \log{\pi_i(O_{i, t}, A_{i, t})} \* \hat{A}_{i,t}^{\textsc{loc}},
\\
L^{\textsc{loc}} &= \lambda_{l}^{\textsc{loc}} \sum_i^{|\mathcal{V}^*|} \sum_t^T{\delta_{i,t}^{\textsc{loc}}} + \sum_i^{|\mathcal{V}^*|} \sum_t^T{\phi_{i,t}^{\textsc{loc}}}\,,
\end{aligned}
\end{equation}
where $r_{i,t}^\textsc{loc}$ denotes the local reward of agent $v_i$ at time step $t$, while $k$ refers to the interval from current state to the next state, following the Semi-POMDP setting in Section \ref{def:semimdp}; $V^{\pi_i}(\cdot)$ is the local value function under the local policy $\pi_i$ of $v_i$; $A_{i, t}^{\textsc{loc}}$ is the action of agent $v_i$; and the hyper-parameter $\lambda_{l}^{\textsc{loc}}$ is the weight of the critic loss. As we can see, the policy loss $\phi_{i,t}^{\textsc{loc}}$ and the critic loss $\delta_{i,t}^{\textsc{loc}}$ are computed based on the advantage function $\hat{A}_{i,t}^{\textsc{loc}}$ derived from the local critic. The training follows the vanilla Actor-Critic algorithm, thus the $\hat{A}_{i,t}^{\textsc{loc}}$ in the third formula are detached in back-propagation and hence will not affect the update of the critic headers.


In the fine-tuning stage, we reinitialize the local actor headers for each agent and create a new global critic header, while the parameters of EncGAT and local critic headers are inherited from the pre-trained part. Different from the pre-training, the global reward $r_t$ is used as a training signal for the local actors' network (including the shared EncGAT and actor header of each agent) and the global critic network, which encourages cooperation towards the global goal. Meanwhile, to prevent the training collapse and avoid the interference of random initialed parameters, we keep using the local critic loss in this stage to supervising the update of the shared EncGAT part with the agent-level local loss derived from local critics. Therefore, the loss function can be formally defined as the linear combination of three parts: the local critic loss $\delta_t^{\textsc{loc}}$, the global critic loss $\delta_t$ and the actor loss $\phi_t$. That is
\begin{equation}
\label{eqn:losses}
\begin{aligned}
\hat{A}_t &= r_{t+1} + \cdots + \gamma^{k-1}r_{t+k} + \gamma^{k}\*{V^{\boldsymbol{\pi}}}\left(S_{t+k}\right) - {V^{\boldsymbol{\pi}}}\left(S_{t}\right),
\\
\delta_t^{\textsc{loc}} &= \sum_i^{|\mathcal{V}^*|} \delta_{t,i}^{\textsc{loc}} = \sum_i^{|\mathcal{V}^*|} {\left(\hat{A}_{i,t}^{\textsc{loc}}\right)}^2, \\
\delta_t &= \hat{A}_t^2, \\
\phi_t &= - \left(\sum_i^{|\mathcal{V}^*|}\log{\pi_i(O_{i, t}, A_{i, t})} \* \right)\hat{A}_t,
\\
L &= \lambda_{l} \sum_t^T{\delta_t^{\textsc{loc}}} + \lambda_{g} \sum_t^T{\delta_t} + \sum_t^T{\phi_t}\,,
\end{aligned}
\end{equation}
where $r_t$ denotes the global reward, $\hat{A}_t$ represents the global state advantage, $V^{\boldsymbol{\pi}}$ is the global value function under the joint policy $\boldsymbol{\pi}$ and the hyper-parameter $\lambda_{l}$ and $\lambda_{g}$ are the weights of the local critic loss and the global critic loss. Similar to the pre-training stage, the gradient of the advantage $\hat{A}_t$ is detached in the training and will not affect the update of the global critic header.

\subsection{Algorithm of EncGAT-PreLAC Framework}
The whole EncGAT-PreLAC framework can be summarized as Algorithm~\ref{alg:fw}. For simplicity of description, we assume all agents belong to the same type while other vertices in the interaction graph can be heterogeneous in the algorithm. Meanwhile, the actor header and critic header are shared among all agents, which can be overwritten according to actual demand. In addition, our framework supports both the synchronous and asynchronous decision of agents, thus
we denote $\mathcal{V}^{*}_t$ as the set of actionable agents at the time step $t$ in both the pre-training and the training procedure.

\begin{algorithm}
\caption{EncGAT-PreLAC Framework}
\label{alg:fw}  
  \begin{algorithmic}[1]
  
  \Procedure{pre-training}{}
  \State Initialize parameters of EncGAT, actor headers and critic headers: $W_e^{\textsc{pre}}, W_{a}^{\textsc{pre}} \text{ and } W_{c}^{\textsc{pre}}$
  \State Initialize the experience pool $D^{\textsc{pre}}=\emptyset$
  \State Initialize simulator $env$
  \For{$k \ \textbf{in} \ 1,\ldots, MAX$}
  \While{$\textbf{not} \ env.done()$}
  \State $\left\{\left(O_{i, t}, \mathcal{A}_{i, t}\right)|i\in \mathcal{V}^{*}_t\right\} \leftarrow env.get\_state()$
  \For {$v_i \in \mathcal{V}^{*}_t$}
  \State $A_{i, t} {\sim} \pi_i \left(A \in \mathcal{A}_{i, t}| O_{i, t};W_e^{\textsc{pre}}, W_{a}^{\textsc{pre}}\right)$
  \EndFor
  \State $\left(r_t, \left\{r_{j,t}^{\textsc{loc}} | j\in \mathcal{V}^{*} \right\}\right) = env.step \left(\left\{A_{i, t}|i\in \mathcal{V}^{*}_t \right\}\right)$
  \State Store experience to $D^{\textsc{pre}}$
  \EndWhile
  \For{batch in $D^{\textsc{pre}}$}
  \State Compute the loss $L^{\textsc{loc}}$ according to formula~\eqref{eqn:loc_losses}
  \State Update $W_{e}^{\textsc{pre}}, W_{a}^{\textsc{pre}}, W_{c}^{\textsc{pre}}$ with the gradient of $L^{\textsc{loc}}$
  \State Reset the simulator $env.reset()$
  \EndFor
  \State Set $D^{\textsc{pre}}=\emptyset$
  \EndFor
  \EndProcedure
  \Procedure{Training}{} 
  \State Initialize the parameters of EncGAT $W_e$ with $W_e^{\textsc{pre}}$
  \State Initialize the parameters of the local critic header  $W_{c}^{\textsc{loc}}$ with $W_{c}^{\textsc{pre}}$
  \State Initialize the parameters of the actor header $W_{a}$ and the global critic header $W_{c}$ with random value
  \State Initialize the experience pool $D=\emptyset$
  \State Initialize simulator $env$
  \For{$k \ \textbf{in} \ 1,\ldots, MAX$}
  \While{$\textbf{not} \ env.done()$}
  \State $\left\{\left(O_{i, t}, \mathcal{A}_{i, t}\right)|i\in \mathcal{V}^{*}_t\right\} \leftarrow env.get\_state()$
  \For {$v_i \in \mathcal{V}^{*}_t$}
  \State $A_{i, t} {\sim} \pi_i \left(A \in \mathcal{A}_{i, t}| O_{i, t};W_e,W_a\right)$
  \EndFor
  \State $\left(r_t, \left\{r_{j,t}^{\textsc{loc}} | j\in \mathcal{V}^{*} \right\}\right) = env.step \left(\left\{A_{i, t}|i\in \mathcal{V}^{*}_t\right\}\right)$
  \State Store experience to $D$
  \EndWhile
  \For{batch in $D$}
  \State Compute the loss $L$ according to formula~\eqref{eqn:losses}
  \State Update $W_e, W_{a}, W_{c}^{\textsc{loc}}, W_c$
  \State $env.reset()$
  \EndFor  
  \State Set $D=\emptyset$
  \EndFor
  \EndProcedure
  
\end{algorithmic} 
\end{algorithm}

\section{Experiments}
\label{sec:exp}
We evaluate our framework on the empty container repositioning problem (ECR) in ocean transportation, which cost billions of dollars per year for shipping companies.

\subsection{Task Description}
\label{sec:task_des}
Empty containers are the core resource used to load goods in ocean transportation. The serious imbalance of import and export among countries and regions all around the world results in extreme imbalance of supply and demand (SnD) for empty containers. With reposition operation, some ports will stack large amount of containers with high storage cost while some other ports have no containers to satisfy customers' shipping requirements. Obviously, both cases will cause serious losses for the shipping companies and hinder the global trade. Empty container repositioning is to leverage the remaining capacity in vessels to explicitly transfer empty containers between deficit ports and surplus ports. Each vessel sails around a route and the travel time is influenced by unpredictable factors such as weather and ocean current. When a vessel arrives at a port, it first loads and discharges cargoes, and then triggers the repositioning action of the port to load to or discharge from the vessel a certain amount of empty containers.

We model the ECR problem as a MAS as follows. We assign an \emph{agent} to each port to manage the containers. When a vessel arrives at the port, the corresponding agent is triggered to make a repositioning decision, which is in fact an event-driven reinforcement learning~\cite{Menda_2019,li2019cooperative}. The repositioning
\emph{action} space is $\{0,1,\cdots, 21\}$ where action 0 means discharging all empty containers on the vessel and action 21 means loading all empty containers of the port on board. Other actions are evenly distributed with a step of 10 percent.
The \emph{observation} of an agent is its own information and that of related ports and vessels.
The optimization goal is to minimize the total shortage of empty containers to fulfill the transportation needs because of the SnD imbalance among ports, which is, on the contrary, to maximize the amount of the fulfilled demand. Therefore, we define the \emph{global reward} as a homogeneous linear function of the number of empty containers used to fulfill transportation orders at a certain time period, that is $r_t = k\cdot \sum_i^n{a_{i,t}}$,
where $k$ is a positive scale factor, $n$ is the number of ports and $a_{i,t}$ is the quantity of consumed containers in port $v_i$ at the time period $t$. 
At the same time, the most straightforward definition of the \emph{local reward} for the port $v_i$ is the value $k\cdot a_{i, t}$.
%

To realize the efficient repositioning, the agents should have the ability to extract useful information from the observation on the heterogeneous interaction graph. As the interactions happen when a vessel arrives at a port, the arrival time of the vessels from different routes plays an important role in decision making. For example, suppose there are two routes, the first one is in lack of containers and the second is rich in containers, namely the demand route and supply route. Two vessels from these routes will arrive at a port in particular time steps. Obviously, whether the vessel in the demand route arrives first or second will substantially affect the optimal repositioning decision of this port, e.g, reserving containers for the incoming vessel or using them out. Such a relative order of two vessels actually poses the requirements for our model to learn the complicate interactions among neighbors, which the vanilla pooling functions like averaging and maximizing will fail to capture as we state in Section~\ref{intro}. 
Another necessary feature for the agents is the cooperation between agents based on agents' comprehensive understanding of the observations. A common phenomenon is that the demand routes and supply routes are not necessarily connected and agents should learn to play as brokers to conduct the multi-hop repositioning.

\subsection{Data and Simulator}
\label{sec:sim}
To build the simulator for the ECR problem, we employ the public routes of an international transportation company, including 22 ports located in main trading countries, 13 routes connecting these ports, and 46 vessels sailing on the routes. The concrete settings of the route topology are attached in the supplementary material. The shipment data between any two ports is generated based on the inter-country trading statistics released by WTO in 2019. 
Because only the yearly statistics are available while the temporal resolution of the simulation is daily, we distribute the total transportation amount into each time-step by two alternative rules. The first is the simple average, the game setting of which is called `normal' and the second is to distribute the amount by a trend function, which is the multiplication of two trigonometric functions with different periods, i.e., 112 time-steps and 28 time-steps, the game setting of which is called `hard'. The transportation time between two ports is computed by their sea distance. At last, noises are added in both transportation time and trading amount to include the environmental uncertainty in each episode. 

\subsection{Input and EncGAT implementation}

\begin{figure}[!ht]
  \centering
  \includegraphics[width=1.0\linewidth,trim= 0mm 0mm 0mm 0mm, clip]{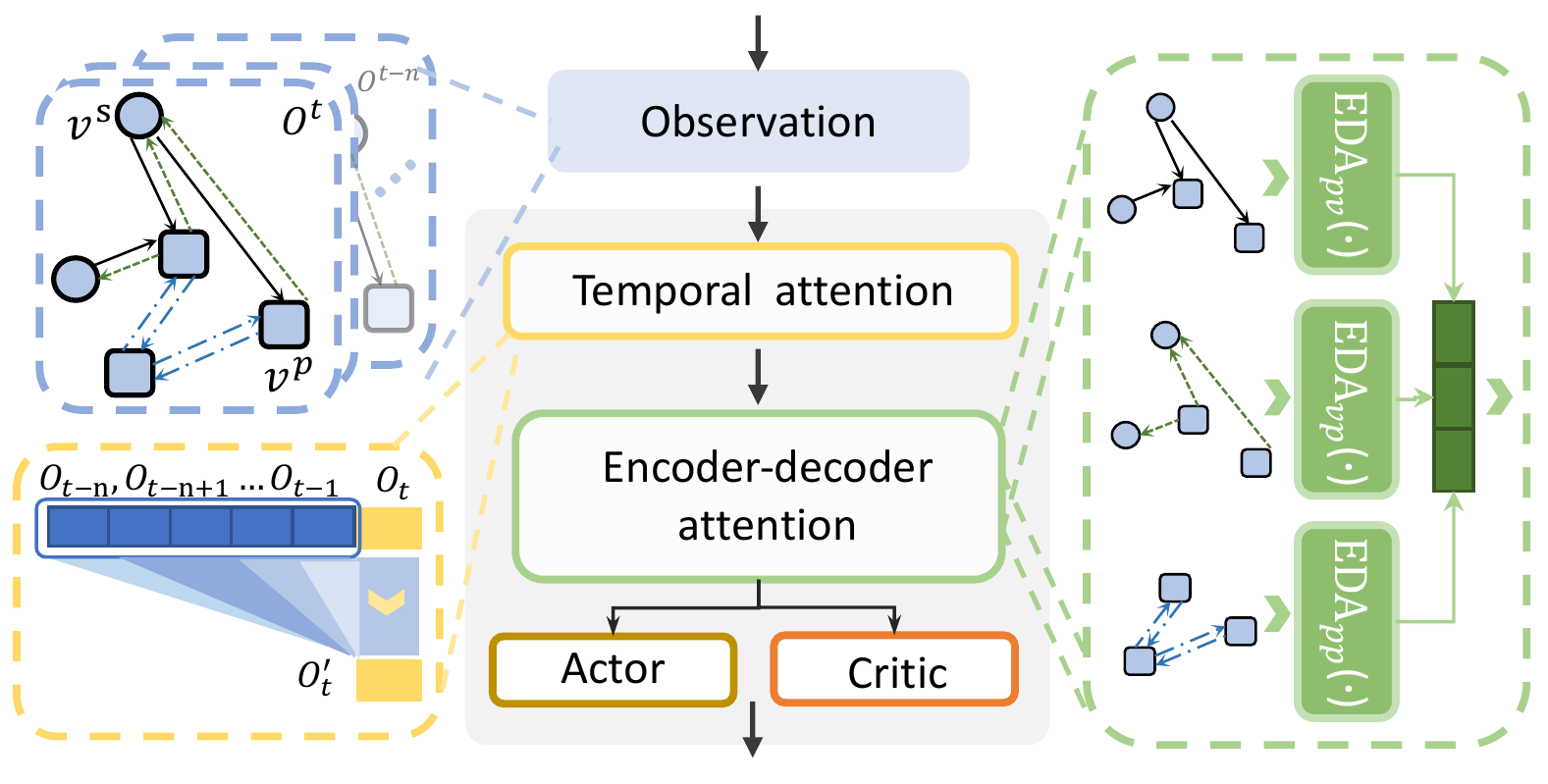}
  \caption{The network structure in ECR task.}
  \label{fig:ecr_network}
\end{figure}
In the ECR problem, there are two kinds of vertices in the interaction graph, ports and vessels denoted as $\left\{\boldsymbol{v}_i^p\right\}$ and $\left\{\boldsymbol{v}_j^s\right\}$ respectively. 
We establish edges from one port to other ports and vessels in the same route, which represents the transportation interactions.
For the port, the input features include the quantities of the empty containers, the reserved empty containers, the laden, the laden on board, transportation orders, the capacity, and the remaining space. The statistical information is also included such as the quantities of the total orders, the fulfilled orders, the failed orders, and their accumulated values in the past 5 time-steps. The input features of the vessel are 7 variables which are the quantities of the empty containers, the laden, the capacity, the remaining space, the accumulated quantities of empty containers, laden, and their difference. As each agent has to learn to predict trend of the inventory and plan for the future, it can hardly determine the strategy with only the input of the current information. Therefore, we let each agent look back a period. We denote the input features of each vertex over the time period $\left[t-n:t\right]$ as $\boldsymbol{x}_{i, \left[t-n:t\right]}$.

The network structure is shown in Figure~\ref{fig:ecr_network}. 
It follows the general design of the proposed framework, but we add three features based on the characteristics of the ECR problem. 
The first one is the temporal attention, which is used to capture the temporal information of the temporally sequential input, is added before the EncGAT module.
For example for vertex $v_i$, the temporal embedding is computed as follows:
\begin{equation}
    \boldsymbol{x}_{i} = \textsc{Att} \left(W^Q_{s}\boldsymbol{x}_{i,t}, W^K_{s}\boldsymbol{x}_{i,\left[t-n:t\right]}, W^V_{s}\boldsymbol{x}_{i,\left[t-n:t\right]}\right)\,\
\end{equation}
where $W_s^{\mathcal{Q}}$, $W_s^{\mathcal{K}}$ and $W_s^{\mathcal{V}}$ are the projection matrices for the query, key and value respectively. In the experiment, the lookback period $n$ is set to 20.
The second one is that to consider the edge feature like the arriving time and the distance, we concatenate it with the features of the corresponding neighbor before feeding it to EncGAT. The third one is that the representations of both the port and the current vessel are fed into the actor header since the vessel situation is important for decision making.

Finally, we stack two encoder-decoder attentions in the EncGAT and residual connections are used to prevent over-smoothing. The actor and critic headers consist of two fully-connected layers also with residual connections. We share the headers of actor and local critic between agents, which makes the overall framework inductive~\citep{NIPS2017_6703}. Empirical studies show the generality of policy with the parameter sharing, which is illustrated in the supplemental file.
 
\subsection{Comparison with Baselines}

{
\begin{table*}[!ht]
\begin{center}
\caption{Fulfillment ratios (\%) of different baselines and ablation experiments. 
}
  \vspace{-2mm}
  \setlength{\tabcolsep}{7mm}
  \label{tab:exp}
  \begin{tabular}{|c|c|c|c|c|}
    \hhline{|-----|}
    \multicolumn{2}{|c|}{\multirow{2}{*}{Section}} & \multirow{2}{*}{Implementation Details} & \multicolumn{2}{|c|}{ Simulation Mode}\\
    \cline{4-5}
    \multicolumn{2}{|c|}{}& & Normal & Hard\\
    \hhline{|-----|}
    \multicolumn{2}{|c|}{\multirow{3}{*}{Baseline}} & No repositioning & $81.95$ & $80.34$ \\
    \multicolumn{2}{|c|}{}& Online LP & $94.00$ & $87.99$\\
    \multicolumn{2}{|c|}{}& DGN in A2C & $90.35 \pm 1.09$ &  $86.50 \pm 0.62$\\
    \cline{1-5}
     & & Normal GC & $85.44 \pm 0.10$ &  $80.77 \pm 0.62$\\
     & {PreLAC} & Separate actors normal GC &  $91.22 \pm 0.12$ & $81.68 \pm 0.05$\\
     {Ablation}& & Separate actors PreLAC & $95.17 \pm 0.29$ & $90.13 \pm 0.40$\\
    \cline{2-5}
     {Study}&\multirow{3}{*}{EncGAT} &Single decoder attention & $98.32 \pm 0.02$ & $90.44 \pm 0.54$\\
     & &GAT & $92.91 \pm 0.68$ & $86.01 \pm 0.09$\\     
     & & LSTM & $86.11 \pm 1.88$ & $80.80 \pm 0.03$\\
    \hhline{|-----|}
    \multicolumn{3}{|c|}{Our Method \quad(EncGAT-PreLAC)} & $\boldsymbol{98.78 \pm 0.21}$ & $\boldsymbol{92.92 \pm 0.38}$ \\
    \hhline{|-----|}
  \end{tabular}
\end{center}
\end{table*}
}

In this section, we compare our framework with existing baseline methods with the total fulfillment ratio, that is the ratio of fulfilled demands to all demands, as the evaluation metric. To keep fairness, we search the learning rate for all deep learning algorithms and fix other hyper-parameters. We run 5 times for each one with the best learning rate. The results are shown in Table~\ref{tab:exp}. It should be pointed out that since most of the ports play the roles of exporter and importer at the same time, the total ratio is around 80 percent if no repositioning is done.

\textbf{Online linear programming (LP)}. In this method, the ECR problem is modeled as an integer linear programming (ILP) problem with mathematical definitions. Here, we use the baseline method in \citep{li2019cooperative}, which applies the rolling horizon policy to do periodic planning. 
It solves the ILP problem for a long period and only takes the first few plans into action, which prevents the deviation of planning from reality. The planning results are largely influenced by the unpredictable noise in the environment, which is the reason why it underperforms our method.
Besides, the online computational cost of searching ILP solutions is much larger than our method.

\textbf{DGN in A2C version}. We chose one of the recent GNN based MARL frameworks called DGN as another baseline and adapt their key contributions into the multi-agent A2C framework, i.e., the network structure and temporal relation regularization. 
DGN directly uses the local reward to train, under the latent assumption that the local reward encourages cooperation, which is not always held in many real-world problems such as ECR. Therefore, we train it with artificially designed local reward, which is a combination of the global and the local reward. As you can see, we also outperform them in both normal and hard mode.

\subsection{Ablation Study}

The ablation studies are used to answer two questions: (1). Does the encoder-decoder attention in EncGAT help heterogeneous information aggregation and intricate interaction understanding? (2). Does PreLAC improve learning performance of cooperative policy?

\textbf{Encoder-Decoder Attention}.
We compare our encoder-decoder attention with alternative aggregation functions, i.e. single decoder attention, GAT~\citep{velivckovic2017graph} and LSTM~\citep{NIPS2017_6703}. 
The single decoder attention directly puts the neighborhood sequence to the attention function in the decoder network introduced in Section~\ref{sec:eda_general}. In the task of the normal mode, its performance is close to our method, but in the task of hard mode, the performance drops by 2\%. The reason is that in the hard mode, the increased noise in the simulation data like the vessels' sailing time and the quantity of orders requires agents to cooperate against uncertainty, making the interactions more complicated.
GAT is different from the single decoder attention in the computation of the attention coefficients. In the original implementation of GAT, the attention coefficients are calculated with a single-layer feedforward network. The result underperforms the dot-product based single decoder attention, which indicates that the way to calculate the attention coefficients may significantly influence the result.
We also try the LSTM aggregator introduced in~\citep{NIPS2017_6703}. Although LSTM also has the ability to consider the inter-correlated influence from neighborhood sequence on an agent, the result is relatively poor. We think the reason is that the recurrent structure makes it difficult to learn with weak signals. 

\begin{figure}[!ht]
  \centering
  \includegraphics[width=0.99\linewidth,trim= 0mm 0mm 0mm 0mm, clip]{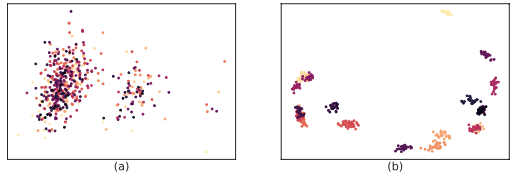}
  \caption{Visualization of vertex embeddings in global-critic (a) and our method (b). The color identifies different ports and each point represents a feature vector in a batch.}
  \label{fig:representation}
\end{figure}

\textbf{Pre-training with local actor-critic}.
To analyze how the pre-training algorithm improves learning performance, we first train the model directly with the global reward (Normal GC). It yields a poor performance especially in the hard mode, which is nearly the same as no repositioning at all. To further understand the result, we visualize the learned embedding of each port by mapping the output vector after EncGAT to the two-dimension space with principal component analysis(PCA). As is shown in Figure~\ref{fig:representation} where the data points of the same port are painted in the same color, we see that with the enhancement of pre-training, the data points regularly gather to several clusters compared to the mass without pre-training in the left picture. This indicates that without the local actor-critic task, the graph layers fail to learn distinguishable vertex embeddings, which will puzzle the succeeding actor and critic headers.
An interesting experiment is also conducted to see the performance if we do not share the actor header between agents (separate actors normal GC), which leads to a little improvement. The reason is that the separate actor headers learn simple but different policies for agents regardless of the poor embedding input. 
Finally, we also try not to share actor headers in our framework (separate actors PreLAC). The result shows about a 2\% drop in performance, which indicates that the parameter sharing, as a space to share experience between agents, helps train a better policy. 

\begin{figure}
  \centering
  \includegraphics[width=0.90\linewidth,trim= 0mm 0mm 0mm 0mm,clip]{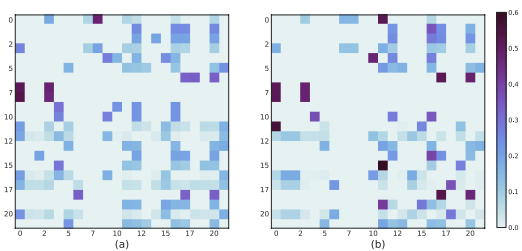}
  \caption{Order proportion distribution between two ports in the original topology (a) and the new topology (b). Each column represents the proportions of orders from one port to orders. Note that two merged ports are the $9^{th}$ and $10^{th}$ columns.}
  \label{fig:order_dist}
  \vspace{-3mm}
\end{figure}

\textbf{Inductiveness and generality}. The inductiveness in the field of GNNs means the ability to generalize to unseen graph structures. Because the EncGAT itself is inductive and we share the actor headers among agents, the policy should be \emph{inductive} too. Henceforth, we further conduct experiments to analyze the generality of the learned policy when the interaction graph changes, e.g., the changes of the topology and the trading distribution in the ECR problem.
In particular, we first merge two ports into one port that takes the roles of the original two ports in their own route. Then, we randomly generate the trading distribution, which is different from the distribution generated from the WTO's data. We plot the heat-map based on the trading quantity in the right of Figure~\ref{fig:order_dist} and compare it with the original one in the left of Figure~\ref{fig:order_dist}. As the distribution completely differs from the original one, the role each port plays and the interactions between agents are different. The normal and hard modes are also defined in this setting. We test the performance of the model trained on the original setting on this new topology \emph{directly}. The result in Table~\ref{tab:tran} shows significant improvements in the fulfillment ratio compared to that with no repositioning.

\begin{table}
\caption{Fulfillment ratios (\%) of the trained model and no repositioning baseline on two test topologies.}
\label{tab:tran}
\setlength{\tabcolsep}{5mm}
\begin{tabular}{c|cc}
\toprule
Mode & Trained model & No repositioning\\
\midrule
Normal & $\boldsymbol{78.69}$ & $67.77$ \\
Hard & $\boldsymbol{76.01}$ & $65.10$ \\
\bottomrule
\end{tabular}
\end{table}

\section{Conclusion and Future Work}
\label{sec:con_fut}
We have proposed a GNN based MARL framework, EncGAT-PreLAC, to solve the emerging challenges in practical cooperative MAS. 
To extract cooperative information from the heterogeneous interaction graph, we leverage the encoder-decoder structure in graph attention to embed heterogeneous features as well as model intricate interactions. To alleviate the training difficulty of GNN under the weak global reward signal, we introduce the pre-training task empowered by the vertex-level local reward. 
Extensive experiments are conducted to solve a practical problem called ECR and the results show the advantages of our method. Nevertheless, some other challenges remain untouched. 
Firstly, the credit assignment problem becomes serious in larger-scaled cooperative tasks. It is interesting to combine techniques like different rewards with our EncGAT for a better assignment. 
Secondly, the interaction information is sometimes unreliable because of either noise or deliberate interference like hacking in real-world problems. How to make cooperation robust to dirty information is practical and promising.
We will conduct our work in these two aspects in the future.

\section{Appendix}
In this section, we will introduce additional experiments and analysis.

\subsection{Source code}
We attach the source code under the folder `\emph{code/}'. The simulator we use is open-sourced, whose link is `\emph{github.com/microsoft/maro}'. All the configurations of experiments are listed in the YAML file `\emph{config.yml}' and all the configurations of the ECR simulation are listed in four YAML files, which are `\emph{normal.yml}', `\emph{hard.yml}', `\emph{shuffle\_normal.yml}' and `\emph{shuffle\_hard.yml}' corresponding to two modes in the original topology and two modes in the 
derived topology. To reproduce the result, you should first download the simulator project from the github, subsitute the configuration file of the ECR simulation, then set the `\emph{model:setting}' field in `\emph{config.yml}' with the experiment name. For example, if you want to reproduce result of the EncGAT-PreLAC framework in the hard setting of the original topology, you should choose the file `\emph{hard.yml}' and change the field `\emph{model:setting}' to `\emph{encgat}'.

\subsection{Settings of the Topology}
\label{sec:topology}
There are $22$ ports located in main trading countries, $13$ routes and $46$ vessels in the transportation topology. The overall topology is shown in Figure~\ref{fig:ecr_network}, where each node is a port and the edges of the same color form a route, whose directions are the sailing order of each vessel in the route. In addition, the two ports painted in blue are the merged ports in the derived topology we introduce in the main text.

\begin{figure}
  \centering
  \includegraphics[width=1.0\linewidth,clip]{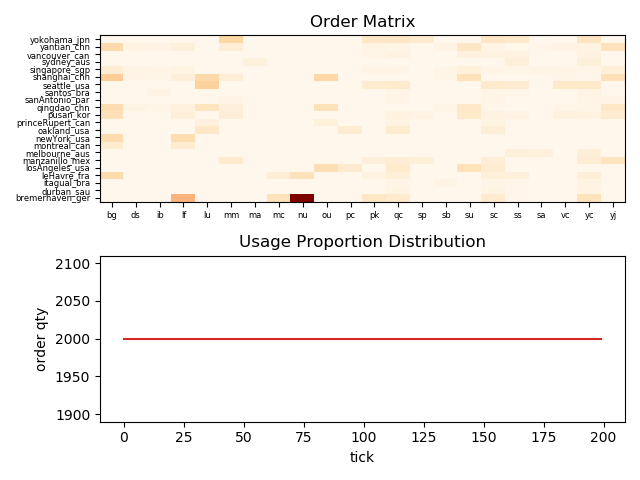}
  \caption{Visualization of the trading distribution in the normal mode. The top is the heat map of transportation proportion from the ports in the row to those in the column; the bottom is the changing curve of total amount of the transportation quantity by the time-step/tick)}
  \label{fig:normal}
  \vspace{-3mm}
\end{figure}

\begin{figure}
  \centering
  \includegraphics[width=1.0\linewidth,clip]{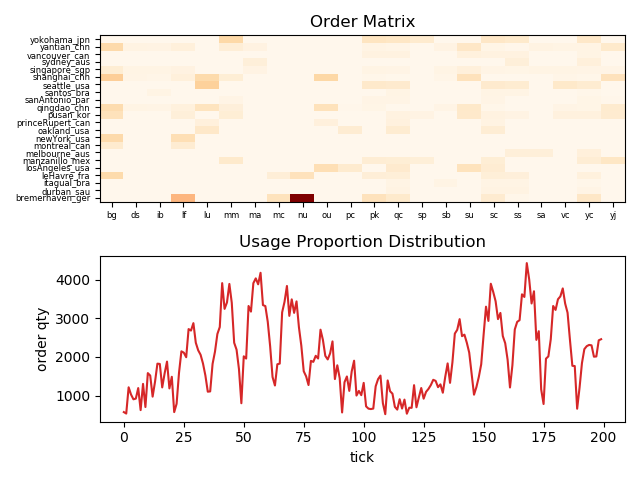}
  \caption{Visualization of the trading distribution in the normal mode. The top is the heat map of transportation proportion from the ports in the row to those in the column; the bottom is the changing curve of total amount of the transportation quantity by the time-step/tick)}
  \label{fig:hard}
  \vspace{-3mm}
\end{figure}

As we stated in the main text, the yearly trading amount are distributed into daily resolution by two rules, e.g., the simple average and the trend function. We plot the trading distribution as well as the total amount of transportation orders at each \emph{time-step} in Figure~\ref{fig:normal} and Figure~\ref{fig:hard}. Suppose the total amount at the time-step $t$ is $\hat{O}_t$, transportation amount between port $v_i$ and port $v_j$ are calculated as a random proportion of $\hat{O}_t$, as is shown in Formula:
\begin{equation}
\label{eqn:order_amt}
\begin{aligned}
&\hat{O}_{t,<i,j>} = \hat{k} * \hat{O}_t\\
&\hat{k} \sim \mathcal{N}(\mu_{<i,j>}, \sigma_{<i,j>})\,,
\end{aligned}
\end{equation}
where $\mathcal{N}$ is the Gaussian distribution and $\mu_{<i,j>}$ and $\sigma_{<i,j>}$ are the constant defined in the configuration file.

\begin{figure}[h!]
  \centering
  \includegraphics[width=1.0\linewidth,,trim= 15mm 0mm 15mm 0mm,clip]{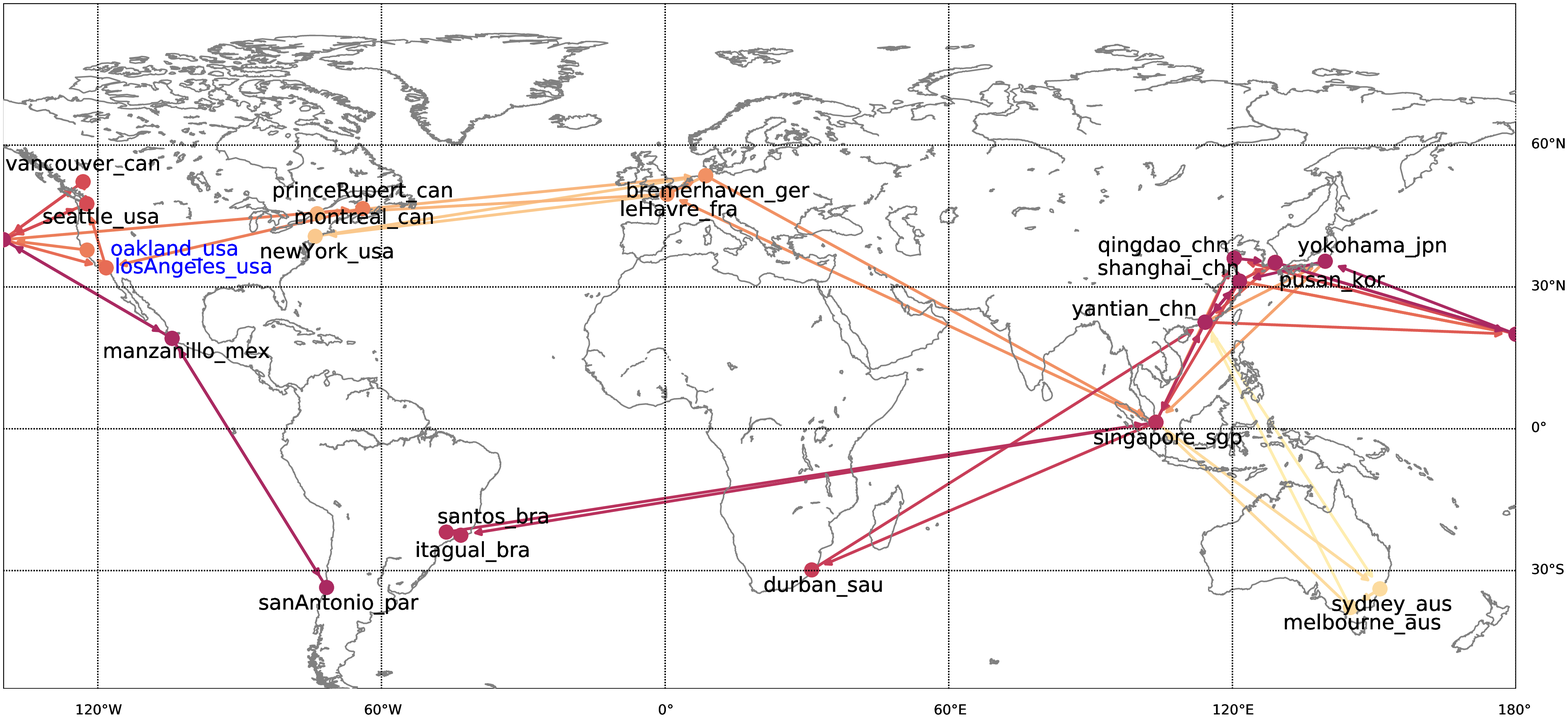}
  \caption{Visualization of the global topology. Each node presents a port and the edges in the same color represents a route.}
  \label{fig:ecr_network}
  \vspace{-3mm}
\end{figure}

\subsection{Experiment Results}
\begin{figure}
  \setlength{\belowcaptionskip}{-3mm}
  \centering
  \includegraphics[width=1.0\linewidth,trim= 0mm 0mm 0mm 0mm,clip]{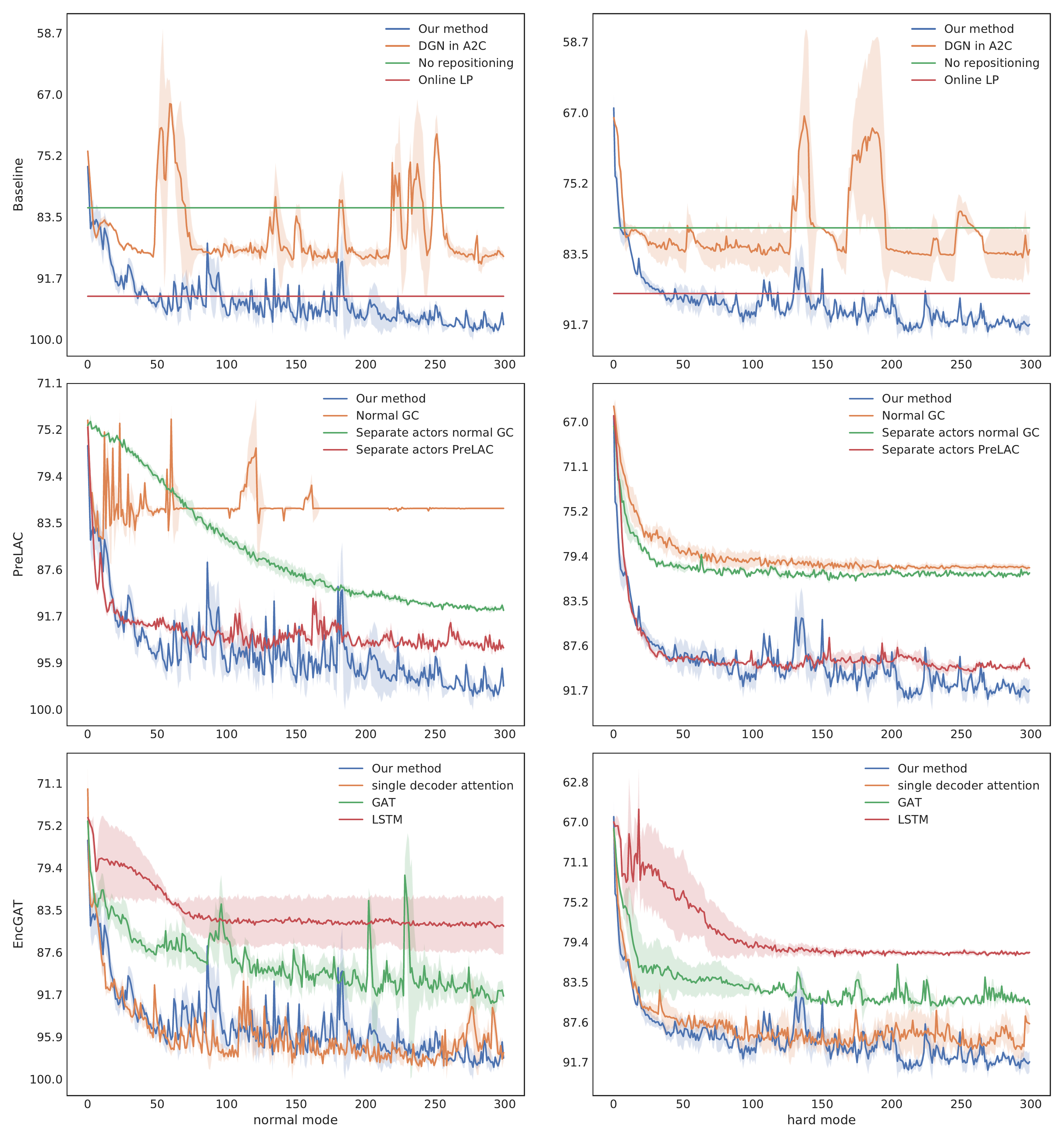}
  \vspace{-2mm}
  \caption{Fulfillment(\%) curves in 300 episode of all the experiments.}
  \vspace{-3mm}
  \label{fig:sh}
\end{figure}

The fulfillment curves of each configuration listed in the result table of the main text are shown in Figure~\ref{fig:sh}.

\clearpage
\bibliographystyle{ACM-Reference-Format} 
\balance
\bibliography{sample}


\end{document}